\documentclass[conference]{IEEEtran}
\IEEEoverridecommandlockouts
\usepackage{cite}
\usepackage{amsmath,amssymb,amsfonts}
\usepackage{algorithmic}
\usepackage{graphicx}
\usepackage{textcomp}
\usepackage{xcolor}
\usepackage{cite}
\usepackage{booktabs}
\usepackage{makecell}
\usepackage{multirow}
\usepackage{graphicx}
\usepackage[draft]{hyperref} 
\usepackage{url}
\def\BibTeX{{\rm B\kern-.05em{\sc i\kern-.025em b}\kern-.08em
    T\kern-.1667em\lower.7ex\hbox{E}\kern-.125emX}}

\makeatletter
\newcommand{\twocolumnfootnotefullwidth}[1]{%
  \begingroup
  \renewcommand{\thefootnote}{}
  \footnotetext{%
    \noindent\hspace*{-1em}\rule{0.3\linewidth}{0.4pt}\\[0.2em] 
    \vspace*{-1em}
    \noindent\footnotesize #1
  }
  \endgroup
}
\makeatother
\begin{document}

\title{COTTA: COntext-aware Transfer adaptation for Trajectory prediction in Autonomous driving\\
}


\author{
\makebox[\textwidth][c]{%
\begin{tabular}{cccc}
Seohyoung Park$^{*}$ & Jaeyeol Lim$^{*}$ & Seoyoung Ju$^{*}$ & Kyeonghun Kim$^{*}$ \\
\textit{Ewha Womans University} & \textit{Seoul National University} & \textit{Sangmyung University} & \textit{OUTTA} \\
03nobel@ewhain.net & limlimlim00@snu.ac.kr & 202115055@sangmyung.ac.kr & kyeonghun.kim@outta.ai
\end{tabular}}%
\\[1.6ex]
\makebox[\textwidth][c]{%
\begin{tabular}{ccc}
Ken Ying-Kai Liao & Nam-Joon Kim\textsuperscript{\dag} & Hyuk-Jae Lee \\
\textit{NVIDIA} & \textit{Seoul National University} & \textit{Seoul National University} \\
kenyingkail@nvidia.com & knj01@snu.ac.kr & hjlee@capp.snu.ac.kr
\end{tabular}}%
}

\maketitle
\twocolumnfootnotefullwidth{* OUTTA affiliation.
\quad {\dag} Corresponding author}

\begin{abstract}
Developing robust models to accurately predict the trajectories of surrounding agents is fundamental to autonomous driving safety. However, most public datasets, such as the Waymo Open Motion Dataset and Argoverse, are collected in Western road environments and do not reflect the unique traffic patterns, infrastructure, and driving behaviors of other regions, including South Korea. This domain discrepancy leads to performance degradation when state-of-the-art models trained on Western data are deployed in different geographic contexts.
In this work, we investigate the adaptability of Query-Centric Trajectory Prediction (QCNet) when transferred from U.S.-based data to Korean road environments. Using a Korean autonomous driving dataset, we compare four training strategies: zero-shot transfer, training from scratch, full fine-tuning, and encoder freezing. Experimental results demonstrate that leveraging pretrained knowledge significantly improves prediction performance. Specifically, selectively fine-tuning the decoder while freezing the encoder yields the best trade-off between accuracy and training efficiency, reducing prediction error by over 66\% compared to training from scratch. This study provides practical insights into effective transfer learning strategies for deploying trajectory prediction models in new geographic domains.
\end{abstract}


\section{Introduction}
With the rapid advancement of autonomous driving technologies, ensuring the safety and reliability of vehicles in complex real-world environments remains one of the most critical challenges. 
Public benchmarks such as the Waymo Open Dataset (Waymo) \cite{waymo} and Argoverse\cite{argoverse,argoverse2} have seen new state-of-the-art (SOTA) models emerge annually.
 
However, despite these advances, a fundamental limitation persists. Most datasets used to train SOTA models are collected in North American or European road environments, reflecting the traffic cultures and infrastructure of those regions. Consequently, they fail to represent the unique characteristics of Asian countries—particularly South Korea—where urban environments are denser, intersections more complex, and driving behaviors often involve aggressive lane changes and frequent cut-ins.

Such dataset bias and domain gaps pose a serious challenge when deploying Western-trained models in domestic contexts, as their prediction accuracy and reliability degrade sharply due to poor domain adaptability. Therefore, it is imperative to develop trajectory prediction models optimized for local road conditions or to establish effective strategies for adapting existing high-performing models.

In this study, we focus on QCNet (Query-Centric Trajectory Prediction)\cite{qcnet}, an innovative query-centric paradigm that has demonstrated both high efficiency and strong predictive performance. QCNet significantly improves computational efficiency during the encoding stage, making it suitable for real-time autonomous driving systems. Nevertheless, since it was pretrained on U.S.-based datasets, it inherits the same domain limitations. Hence, our objective is to systematically analyze how effectively a pretrained QCNet can adapt to the congested and dynamic road environments of South Korea.

To address this, we design and compare four distinct transfer-learning strategies using a Korean autonomous-driving dataset. Through this comparative analysis, we evaluate the adaptability of QCNet to local driving conditions and propose an adaptation strategy that achieves both optimal performance and training efficiency. The findings of this research are expected to provide practical guidelines for successfully integrating and localizing state-of-the-art trajectory prediction models in future Korean autonomous driving systems.


\begin{figure*}[t]
  \centering
\includegraphics[width=0.9\textwidth]{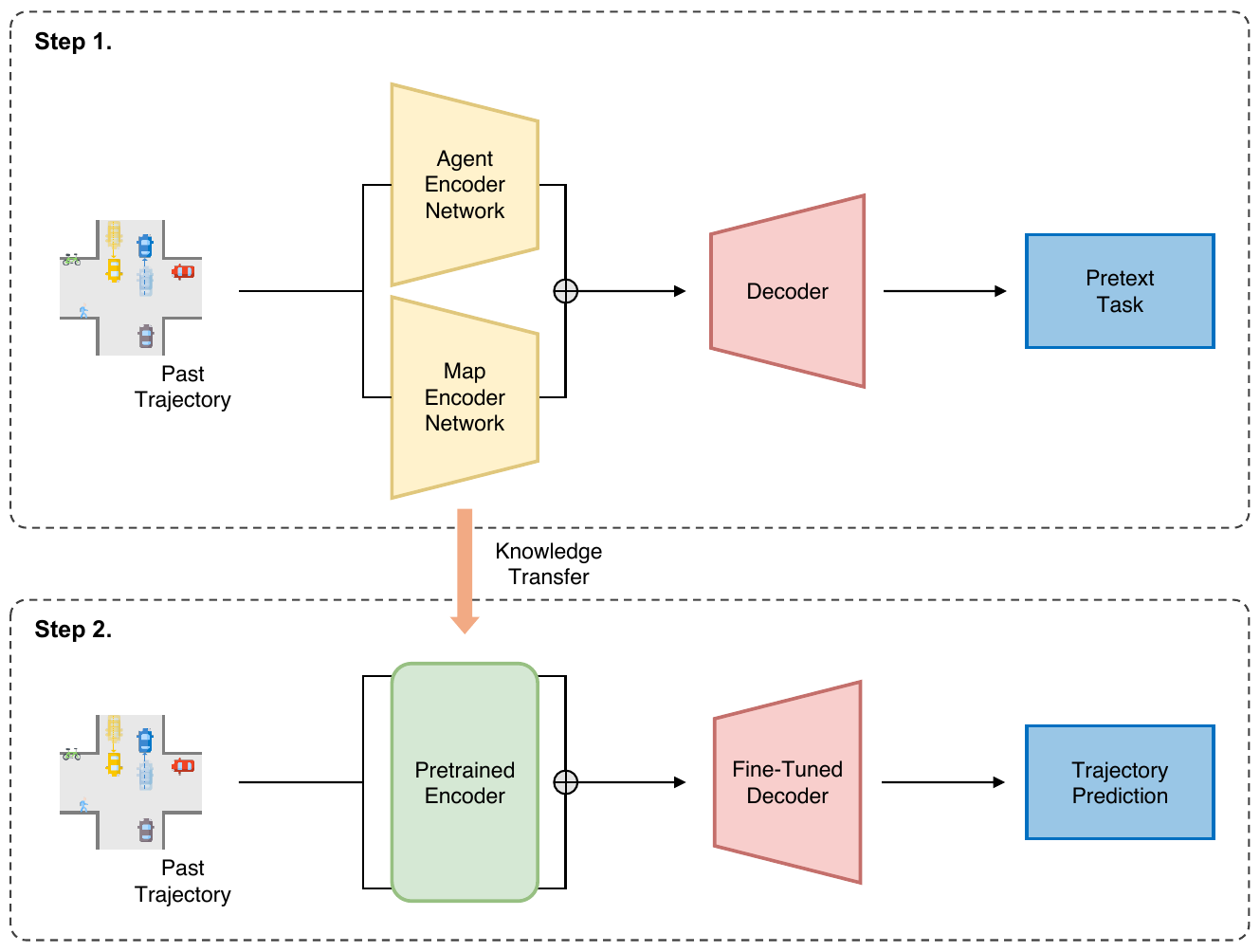}
  \caption{Overview of the proposed methodology}
  \label{fig:fig1}
\end{figure*}

\section{RELATED WORKS}
\subsection{Motion Forecasting}\label{motion_forecasting}
QCNet\cite{qcnet}, introduced at CVPR 2023, marks the origin of the query-centric paradigm in trajectory prediction and has become a milestone model in autonomous driving research.
By reformulating agent–scene interactions through query-based reasoning, QCNet achieved both high efficiency and accuracy, ranking first on the Argoverse 1 and 2 benchmarks, and winning the CVPR 2023 Workshop on Autonomous Driving  Challenge.

However, the Argoverse  dataset used to train QCNet was primarily represents U.S. urban driving environments (e.g., Pittsburgh, Miami), reflecting Western traffic patterns and infrastructure, which potentially limits the model's adaptability to non-Western contexts such as densely populated Asian cities. For instance, UniTraj\cite{unitraj} introduced a unified benchmark across Waymo\cite{waymo}, Argoverse 2\cite{argoverse2}, and nuScenes \cite{nuscenes}, conducting cross-dataset evaluations that revealed a substantial performance drop when models trained on one dataset were tested on another. Similarly, models trained on Argoverse 2 show a 20–30\% degradation in b-minFDE6 when tested on nuScenes, underscoring how dataset diversity and scale fundamentally affect generalization performance \cite{goal}.

We further hypothesize that country-specific road structures and driving behaviors also contribute to the domain gap. Consequently, a model like QCNet, trained on U.S. datasets, is expected to experience performance degradation when applied to Korean driving environments.
In this work, we systematically analyze this phenomenon by evaluating QCNet on Korean autonomous-driving datasets and investigate effective transfer-learning strategies to enhance domain adaptability and prediction performance in region-specific contexts.

\section{EXPERIMENTS}
\subsection{Datasets}\label{dataset}
The Argoverse \cite{argoverse,argoverse2} comprises a large-scale collection of real-world driving data captured from autonomous vehicles in diverse urban environments. The subsequent release, Argoverse 2 motion forecasting dataset, significantly enhances its predecessor by providing 250,000 scenarios that span more than 2,000 km of roadways across six geographically diverse cities. Each scenario contains 11 seconds of tracking data accompanied by high-definition vector maps that include lane centerlines and road topology information. The dataset partitions each scenario into a 5-second observation period followed by a 6-second prediction horizon, enabling comprehensive evaluation of motion forecasting algorithms in complex multi-agent environments.

We used the ETRI Trajectory Prediction Challenge 2025 dataset \cite{etri}. The ETRI Trajectory Dataset (ETD) contains 20,739 scenarios collected by two vehicles following predefined routes in Daejeon and Sejong, South Korea. The dataset includes trajectories of vehicles, cyclists, and pedestrians. Data samples are provided as 6-second segments, comprising 2 seconds of historical data and 4 seconds of future data. In addition, the dataset provides an official split for trajectory prediction. The training set is generated from approximately 28,800 collected frames, creating learning scenes that contain roughly 494,000 training trajectories. The evaluation set is generated from approximately 4,200 collected frames, creating test scenes that contain roughly 84,000 test trajectories. For evaluation, we focus exclusively on vehicle trajectories.

\begin{figure}[t]
  \centering
  \includegraphics[width=\columnwidth]{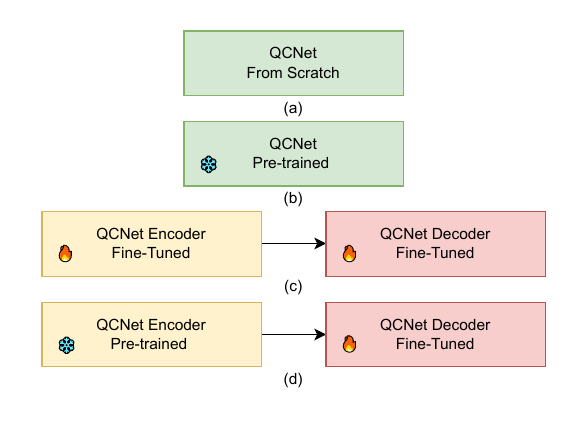}
  \caption{Overview of transfer learning strategies from Argoverse 2 to ETRI Trajectory Dataset: (a) Zero-shot evaluation, (b) Training from scratch, (c) Full fine-tuning, (d) Encoder freezing fine-tuning.}
  \label{fig:fig2}
\end{figure}

\subsection{Experimental Details}\label{details}
All experiments were conducted on an NVIDIA A100 GPU for 150 epochs with batch size 16 for training and validation. Map elements were filtered to those within 50 m of the target agent. The models were trained using the AdamW optimizer \cite{adamw} with cosine annealing, weight decay of $1e^{-4}$, dropout of 10\%, and learning rate of $1e^{-4}$.

\subsection{Experimental Setup}\label{setup}
To investigate the domain gap that arises when applying a QCNet model pretrained on the U.S.-based Argoverse 2 to the ETRI Trajectory Dataset, we conducted four transfer learning experiments designed to assess performance degradation and identify an effective adaptation strategy. For all experiments, we utilized the official QCNet implementation with the pretrained model weights released by the authors. The four experimental settings are as follows:
\begin{itemize}

\item \textbf{Zero-shot}: The QCNet model pretrained on Argoverse 2 was directly evaluated on the ETRI test set without any further training.
\item \textbf{Scratch Training}: The QCNet architecture was retained, but the model was trained from scratch using only the ETRI training set, and subsequently evaluated on the ETRI test set.
\item \textbf{Full Fine-tuning}: The entire QCNet model pretrained on Argoverse 2 was fine-tuned using the ETRI training set and then evaluated on the ETRI test set.
\item \textbf{Encoder Freezing}: The QCNet model pretrained on Argoverse 2 was adapted by freezing the encoder and fine-tuning only the decoder with the ETRI training set, followed by evaluation on the ETRI test set.
\end{itemize}

\subsection{Metrics}\label{metric}

\begin{equation}
minADE_K = \frac{\min_{l=1}^{K} \sum_{t=T_{\text{obs}}}^{T_{\text{pred}}-1} {\left\| \hat{y}_i^{t,(l)} - y_i^t \right\|}_2}{T_{\text{pred}} - T_{\text{obs}}},\label{ade}
\end{equation}

\begin{equation}
\min FDE_K = \min_{l=1}^{K} {\left\lVert \hat{y}_i^{T_{\text{pred}-1,(l)}} - y_i^{T_{\text{pred}-1}} \right\rVert}_2,\label{fde}
\end{equation}
The evaluation metrics were minimum Average Displacement Error (minADE) and minimum Final Displacement Error (minFDE), as defined in \eqref{ade} and \eqref{fde}.

\begin{equation}
NLL_{\text{Laplace}} = \sum_{t=T_{\text{obs}}}^{T_{\text{pred}}-1} \log P(y_j^t | \hat{y}_j^t, \hat{b}_j^t),\label{laplace}
\end{equation}

\begin{equation}
\mathcal{L}_{\text{total}} = \mathcal{L}_{\text{propose}} + \mathcal{L}_{\text{refine}} + \mathcal{L}_{\text{cls}},\label{loss}
\end{equation}

The training procedure followed the original QCNet \cite{qcnet} training configuration, and the loss functions in \eqref{laplace} and \eqref{loss} were applied accordingly during optimization.

\section{RESULTS}
We evaluate the transferability of QCNet, pretrained on Argoverse 2, to the Korean driving domain. The quantitative results of the four training strategies are presented in Table~\ref{tab:strategy_comparison}, comparing minADE and minFDE performance. The zero-shot results (Pretrained) confirm a clear domain gap, as the performance significantly degrades when a model trained on U.S. driving data is directly applied to Korean road scenarios. Training QCNet from scratch on the ETRI dataset reduces minADE but leads to a substantial increase in minFDE, indicating difficulty in learning long-term motion patterns due to limited data scale. By contrast, transfer learning strategies that leverage pretrained weights show notably superior performance. 

In particular, the Encoder Freezing approach achieves the best results, with a 66.3\% reduction in minADE and an 83.7\% reduction in minFDE compared to Scratch Training. This demonstrates that large-scale pretraining provides essential general motion representations, while fine-tuning with Korean data enables the model to effectively learn domain-specific driving behaviors and traffic patterns.

Overall, the results validate that pretraining + targeted fine-tuning is crucial for adapting trajectory prediction models to new geographic domains, and that selectively updating the decoder offers an efficient and highly effective transfer strategy for localizing QCNet to Korean road environments.

\section{Conclusion}
This study explored the optimal transfer learning strategy for adapting QCNet—an advanced trajectory prediction model pretrained on the U.S.-based Argoverse 2 dataset—to the high-density and highly interactive driving environments of South Korea. Among the four evaluated strategies, the Encoder Freezing approach achieved the best localization performance, reducing minADE by approximately 66.3\% compared to Scratch Training. 

These findings confirm that large-scale pretraining provides strong general motion priors, and that selectively fine-tuning model components enables efficient acquisition of domain-specific driving behaviors. The results highlight the importance of positive transfer from source domains and demonstrate that the proposed strategy is a practical and effective pathway for localizing SOTA trajectory models for new geographic regions.

While the transfer learning approaches demonstrated effectiveness, qualitative analysis reveals that residual domain discrepancies persist due to the unique characteristics of Korean road environments. To address these limitations, future work will pursue the following research directions:
\begin{itemize}

\begin{table}[t]
\centering
\caption{PERFORMANCE COMPARISON OF QCNET TRAINING STRATEGIES} 
\label{tab:strategy_comparison}
\begin{tabular}{lcc}
\hline
\textbf{Training Strategy} & \textbf{minADE} & \textbf{minFDE} \\
\hline
Pretrained & 0.767 & 1.460 \\
Scratch Training & 0.540 & 2.720 \\
Full Fine-tuning & 0.205 & 0.721 \\
Encoder Freezing & \textbf{0.182} & \textbf{0.444} \\
\hline
\end{tabular}
\end{table}

\item \textbf{Data-Centric Enhancement}:
We will strengthen the representation of Korean road characteristics by improving both the quality and diversity of training data. This includes expanding domestic data collection to cover a broader range of geographic areas and complex scenarios, as well as employing targeted data augmentation techniques that simulate region-specific behaviors, such as high-density interactions and aggressive lane changes. These efforts aim to enhance model robustness in diverse real-world settings.

\item \textbf{Domain Adaptation Techniques}:
Beyond supervised fine-tuning, we will adopt Unsupervised Domain Adaptation (UDA) methods to explicitly reduce distributional shifts between source and target domains. Approaches such as adversarial learning, feature alignment, and style transfer will be explored, particularly applied to the encoder to reduce structural and contextual feature discrepancies between Argoverse-like and Korean road data.

\item \textbf{Multi-Domain Learning Framework}:
To expand adaptability across multiple countries, we plan to investigate Multi-Domain Learning strategies that enable a single model to generalize across heterogeneous road cultures. Techniques such as domain discriminators, domain-specific encoders, and mixture-of-experts architectures will be employed to jointly preserve global generalization while retaining local specialization.
\end{itemize}

In addition, future work may explore online adaptation, where the model continuously refines itself using in-field driving data, allowing real-time adaptation to evolving traffic patterns. We also foresee value in extending this study to incorporate social interaction modeling and map-less prediction architectures, enabling robust performance in regions with limited or outdated HD maps.

\section*{Acknowledgment}
This study was supported by the Next Generation Semiconductor Convergence and Open Sharing System, as well as by the Institute of Information \& Communications Technology Planning \& Evaluation (IITP) within the Artificial Intelligence Semiconductor Support Program to Nurture the Best Talents (IITP-2023-RS-2023-00256081), managed by the Korea government (MSIT).

\vspace{12pt} 

\bibliographystyle{IEEEtran}
\bibliography{references}

\end{document}